\documentclass[letterpaper]{article} % DO NOT CHANGE THIS
\usepackage[]{aaai2026}  % DO NOT CHANGE THIS
\usepackage{times}  % DO NOT CHANGE THIS
\usepackage{helvet}  % DO NOT CHANGE THIS
\usepackage{courier}  % DO NOT CHANGE THIS
\usepackage[hyphens]{url}  % DO NOT CHANGE THIS
\usepackage{graphicx} % DO NOT CHANGE THIS
\usepackage{natbib}  % DO NOT CHANGE THIS AND DO NOT ADD ANY OPTIONS TO IT
\usepackage{caption} % DO NOT CHANGE THIS AND DO NOT ADD ANY OPTIONS TO IT
\usepackage{algorithm}
\usepackage{algorithmic}

\usepackage{newfloat}
\usepackage{listings}
\DeclareCaptionStyle{ruled}{labelfont=normalfont,labelsep=colon,strut=off} % DO NOT CHANGE THIS
\floatstyle{ruled}
\newfloat{listing}{tb}{lst}{}
\floatname{listing}{Listing}
\nocopyright %-- Your paper will not be published if you use this command
\usepackage{booktabs}
\usepackage{amsfonts}
\usepackage{amsmath}
\usepackage{pgfplots}
\pgfplotsset{compat=1.18}

\title{LoRA-Augmented Generation (LAG) \\ for Knowledge-Intensive Language Tasks}
\author{
    William Fleshman,
    Benjamin Van Durme
}
\affiliations{
    Johns Hopkins University\\
    3400 N. Charles Street \\
    Baltimore, MD 21218 USA\\
    will.fleshman@jhu.edu,
    vandurme@jhu.edu
}

\begin{document}

\maketitle

\begin{abstract}
The proliferation of fine-tuned language model experts for specific tasks and domains signals the need for efficient selection and combination methods. We propose LoRA-Augmented Generation (LAG) for leveraging large libraries of knowledge and task-specific LoRA adapters. LAG requires no additional training or access to data, and efficiently filters, retrieves, and applies experts on a per-token and layer basis. We evaluate LAG on various knowledge-intensive tasks, achieving superior performance over existing data-free methods. We explore scenarios where additional data is available, demonstrating LAG's compatibility with alternative solutions such as retrieval-augmented generation (RAG). 
\end{abstract}

\begin{links}
    % \link{Code}{https://aaai.org/example/code}
\end{links}

\section{Introduction}

Modern language models (LMs) pretrained on large general-purpose text collections have led to rapid gains in task performance. Significant research is now focused on methods for for effectively deploying them in novel data domains and for specialized applications \citep{rag, lora, taskvector, readapt, scpt}. One prominent approach involves leveraging techniques such as Retrieval-Augmented Generation (RAG), which dynamically injects relevant information from a collection of documents at inference time to guide the model output \citep{rag}. While highly effective, RAG necessitates the availability and efficient retrieval of these documents during the inference phase.

Alternatively, LMs can be adapted to specific tasks or custom data domains through fine-tuning, a process that adjusts the model parameters using task-specific or domain-specific datasets \citep{bapna-firat-2019-simple, adapters, Wei2021FinetunedLM, peft}. A particularly efficient and popular fine-tuning method is Low-Rank Adaptation (LoRA), which introduces a small number of trainable parameters, known as LoRA adapters, alongside the frozen pretrained model \citep{lora}. This approach significantly reduces computational costs and storage requirements compared to full fine-tuning. The success and efficiency of LoRA has led to a rapid proliferation of these adapters, with numerous variations and specialized versions openly shared and readily accessible in public repositories such as Hugging Face \citep{huggingface}. The abundance of these readily available LoRA adapters, each potentially specialized for different tasks, domains, or styles, presents a unique opportunity. However, it also introduces a critical challenge: how to effectively select or combine these diverse adapters at inference time to achieve optimal performance. 

Many existing methods for leveraging LoRA adapters rely on access to the original training data corresponding to each adapter or require additional training to merge or select them \citep{pfeiffer-etal-2021-adapterfusion,wang-etal-2022-adamix, caccia2023multihead, ponti-etal-2023-combining, adapterswap, lorahub,zadouri2024pushing}. For example, \textit{Parametric-RAG (PRAG)} uses a RAG-like retrieval mechanism over training documents, but loads an adapter corresponding to the selected document instead of including the content of the document in the prompt \citep{prag}. The dependency on data or additional training can be a significant bottleneck, especially in scenarios where the data is proprietary, unavailable, or too large to manage. These issues have led to the recent development of unsupervised routing methods. Arrow routing constructs rank-1 prototypes directly from the LoRA weights and uses them to efficiently select adapters on-the-fly\,\citep{arrow}. Spectral routing (SpectR) improves the accuracy of Arrow at the cost of higher computational complexity \citep{spectr}.

\begin{figure*}[t]
    \centering
\includegraphics[width=2\columnwidth]{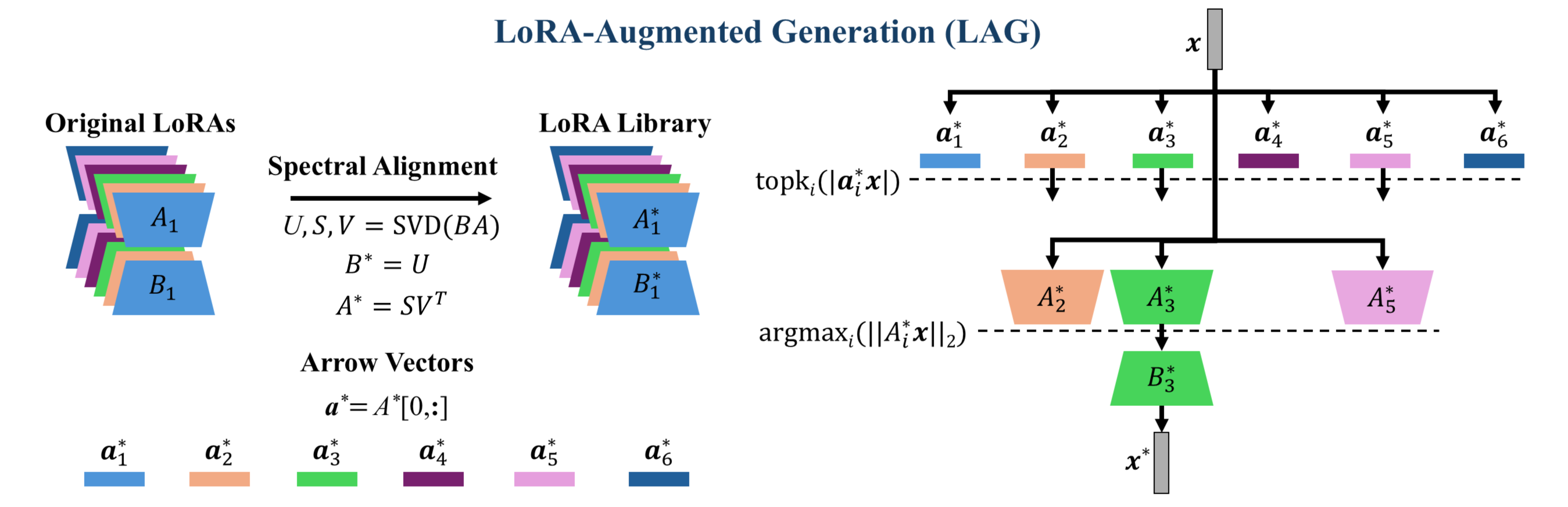}
    \caption{Overview of LAG. LoRA adapters are converted offline via SVD to align representations and extract \textit{arrows}. The token vector $\mathbf{x}$ is processed in two stages: (1) \emph{Arrow routing} is used to efficiently filter the large library of adapters to a smaller set of k potential LoRAs, and (2) \emph{Spectral routing} is used to rank the filtered selection by measuring the length of the token representation in the basis of each adapter. The best adapter completes the new token representation $\mathbf{x}^*$.}
    \label{lag}
\end{figure*}

We introduce LoRA-Augmented Generation (LAG), an approach developed to address the challenge of effectively utilizing existing LoRA adapters without requiring their corresponding data or any additional training. LAG delivers a flexible and efficient mechanism for leveraging the collective knowledge and capabilities embedded within a large set of LoRA adapters, enabling dynamic LoRA selection on a per-token and per-layer basis (Figure \ref{lag}). Specifically we:
\begin{itemize}
    \item Develop an efficient LoRA selection and routing procedure to outperform other training-free approaches.  
    \item Leverage a library of over 1000 LoRAs to demonstrate our improved results on knowledge-intensive tasks; and
    \item Compare and combine LAG with alternative methods for augmenting LMs with additional knowledge.
\end{itemize}

\section{Background}
This section provides an overview of the key concepts and prior works motivating our problem setting and proposed LAG framework. We discuss existing paradigms for incorporating new knowledge into language models, focusing on retrieval-augmented generation and parameter-efficient fine-tuning methods, particularly LoRA. We then delve into recent advancements in knowledge acquisition, adapter retrieval, and unsupervised routing techniques, highlighting the strengths and limitations of each approach. 

\subsection{Retrieval-Augmented Generation}
Retrieval-Augmented Generation (RAG) has emerged as a powerful paradigm for grounding LMs with external knowledge, thereby mitigating issues like hallucination and enabling access to up-to-date information \citep{rag}. The core idea behind RAG is to augment the LM's input with relevant passages retrieved from a knowledge base. A typical RAG pipeline involves several stages: a retriever component fetches relevant documents or passages based on a query, an optional reranker then reorders these retrieved passages to select the most pertinent ones, and finally, these selected passages are concatenated with the user query and fed into the generative model \citep{glass-etal-2022-re2g}. The retriever can be based on sparse methods like BM25 or dense methods using embedding models \citep{bm25, karpukhin-etal-2020-dense, gao2024retrievalaugmented}. While highly effective, RAG systems face several challenges. Managing large and dynamic knowledge bases, ensuring retrieval relevance for diverse queries, and fitting retrieved context within the LM's finite context window are all significant hurdles \citep{Liu2023LostIT, gao2024retrievalaugmented, sevenrag}. Moreover, the performance of a RAG solution heavily depends on the quality and freshness of the underlying knowledge base, making robust RAG difficult to implement, especially in rapidly evolving domains where the dataset itself is dynamic and constantly updated. Critically, RAG solutions require the external data to be available and retrieved at inference time, which might not always be feasible. In this work, we explore the case where knowledge is only available parametrically via adapters, but also compare to scenarios where relevant documents can be retrieved via RAG. 

\subsection{Parameter-Efficient Fine-Tuning (PEFT)}
Beyond RAG, fine-tuning is another prominent approach to adapt LMs to new tasks or domains. However, full fine-tuning of large models is computationally expensive and memory-intensive, leading to the development of Parameter-Efficient Fine-Tuning (PEFT) methods \citep{peft}. PEFT techniques aim to update only a small subset of model parameters while keeping the majority of the pretrained weights frozen, significantly reducing computational cost and storage. Popular PEFT methods include prompt tuning, prefix tuning, and adapter-based approaches \citep{bapna-firat-2019-simple, adapters, lester-etal-2021-power, li-liang-2021-prefix}.

Among these, \citet{lora}'s Low-Rank Adaptation (LoRA) has gained significant traction due to its effectiveness and simplicity. LoRA fine-tunes LMs by injecting trainable low-rank decomposition matrices into the transformer layers. Specifically, for a weight matrix $W \in \mathbb{R}^{m \times n}$, LoRA introduces two smaller matrices $A \in \mathbb{R}^{r \times n}$ and $B \in \mathbb{R}^{m \times r}$ where $r \ll \min(m,n)$ is the LoRA rank such that the update to the original weight matrix is represented as $W + BA$. During training, $W$ is frozen, and only $A$ and $B$ are optimized. At inference, the full matrix $W+BA$ can be computed and used, or $A$ and $B$ can be dynamically loaded and applied \citep{peft}. This approach dramatically reduces the number of trainable parameters, enables faster training, and allows for the easy storage and swapping of multiple adapters for different tasks or knowledge domains without requiring modifications to the base model \citep{lora, chronopoulou-etal-2023-adaptersoup, adapterswap}. \textit{Task-vectors} isolate the difference in weights regardless of fine-tuning strategy \citep{taskvector}, and \citet{readapt} demonstrates that the SVD can be used to convert these differences into LoRA-like adapters, enabling broader applicability for our LoRA-specific approach.

\subsection{Knowledge Acquisition beyond RAG}
While RAG augments models with external information at inference, several methods have been developed to embed new knowledge into existing language models \citep{lu-etal-2021-parameter-efficient, zhang-etal-2023-plug, kblam, scpt, km}. These approaches often involve training the model to internalize specific facts or domains. For instance, \textit{Synthetic continued pretraining} constructs a knowledge graph from target data and uses LMs to generate additional training data by sampling relationships from the graph \citep{scpt}. \citet{physics} find that data augmentation is essential for LMs to extract knowledge, a finding which adapter-based methods have also leveraged \citep{km, prag}. 

\subsection{Adapter Retrieval}
The proliferation of LoRA adapters necessitates effective methods for selecting or combining them at inference time, especially when dealing with a vast library of adapters each specialized for a particular task or knowledge domain. Recent works leverage data associated with adapters to aid in retrieval \citep{zhao-etal-2024-loraretriever, prag}. \textit{Parametric-RAG (PRAG)} proposes selecting knowledge adapters based on the similarity between a query and the training data used for each adapter \citep{prag}. They employ BM25 over training documents and match the selected document to the associated adapter. Like traditional RAG, this approach can be effective for conditioning models on external knowledge if the corresponding data is available. Similarly, \textit{LoRARetriever} constructs task representations using examples from the training data of each adapter and then trains a retriever to select the most appropriate task adapter for a given query \citep{zhao-etal-2024-loraretriever}. While promising, both PRAG and LoRARetriever leverage the associated training data. In this work, we focus on data- and training-free adapter retreival. 

\subsection{Unsupervised Adapter Routing}
To overcome the data and training dependencies of adapter retrieval methods, recent research has explored unsupervised routing techniques that leverage the inherent properties of the adapter weights themselves \citep{arrow, spectr}. \textit{Arrow routing} is an efficient method that projects adapter weights into a single principal component. The routing decision is then made based on the similarity of the query embedding to these one-dimensional projections. While computationally efficient, relying on a single dimension can lead to inaccuracies due to the inherent loss of information from higher dimensions. \textit{Spectral routing (SpectR)} addresses this issue by utilizing the full spectral properties of the adapter weights, providing a more accurate representation for routing decisions \citep{spectr}. However, this increased accuracy comes at the cost of higher computational complexity, as it involves working with the full rank of the adapters. 
% Our proposed solution leverages the complementary strengths of these two methods to achieve efficient performance.
This extra cost is compounded by the number of available adapters, making SpectR untenable with large LoRA libraries. LAG leverages the complementary strengths of these two methods, yielding efficient performance with LoRA libraries too large for SpectR processing.

\subsection{Problem Setting}
The number of pretrained LMs is increasing, and so to are repositories of specialized LoRA adapters. These are often fine-tuned for specific tasks or imbued with domain-specific knowledge. This motivates a need for methods to select and apply these adapters at inference time. Specifically, we explore the setting where a large library of \textit{knowledge adapters} is available, each adapter trained on a specific body of knowledge. Likewise, we assume a library of \textit{task-adapters}, where each adapter is trained for a knowledge-intensive task. Our core objective is to develop an approach for selecting and applying the knowledge and task adapters most relevant to a specific query without access to the corresponding data or additional training for learning to route. 

\section{LoRA-Augmented Generation}

Our method introduces an efficient framework for leveraging large libraries of existing LoRA adapters during LM inference, which we term LoRA-Augmented Generation (LAG). We assume access to two distinct sets of LoRA adapters trained for different purposes: (1) a task library $\mathcal{T}$ comprising adapters specialized for specific tasks (fact checking, question answering, etc.), and (2) a knowledge adapter library $\mathcal{K}$ containing LoRAs that encode domain or document-specific information (e.g., Wikipedia articles).

To perform inference with a base LM augmented by these adapter libraries, we propose a two-stage routing strategy for dynamically selecting adapters on a per-token and per-layer basis. Our approach addresses the challenge of scaling inference to thousands of available adapters while maintaining computational efficiency and task-specific performance.

\subsection{Spectral Alignment}

First, we perform offline processing of our adapter libraries following \citet{spectr}'s \textit{spectral alignment} procedure. For LoRA weight matrices $A$ and $B$ with rank $r$, we calculate the rank-$r$ singular value decomposition (SVD) of the matrix product $BA$:
\begin{equation}
    U,S,V = \mathrm{SVD}_r(BA),
\end{equation}
with $U \in \mathbb{R}^{m \times r}$ and $V \in \mathbb{R}^{n \times r}$ the left and right singular vectors and $S \in \mathbb{R}^{r \times r}$ the diagonal matrix of singular values such that:
\begin{equation}
    USV^T = BA.
\end{equation}
The adapter is stored as new matrices $B^*$ and $A^*$ where:
\begin{equation}
    B^* = U, \text{ and }
\end{equation}
\begin{equation}
     A^* = SV^T.
     \label{Astar}
\end{equation}
Importantly, the SVD is performed once offline and results in an adapter of equivalent size and function. \citet{arrow} store adapters in their original form but use $\mathbf{a}^*$, the row vector from $A^*$ associated with the largest singular value, as the adapter prototype for their Arrow routing algorithm. The matrix $A^*$ contains the scaled eigenvectors of the covariance matrix of the LoRA parameters, which represent orthogonal directions of maximum variation induced by the adapter in the space of input vectors $\mathbf{x} \in \mathbb{R}^{n}$ \citep{arrow, spectr}. The vector $\mathbf{a}^*$ is the direction capturing the most variation and produces the largest adapter activations of all unit-length input vectors:
\begin{equation}
    \mathbf{a}^* = \mathrm{argmax}_{\mathbf{x},\lvert\lvert\mathbf{x}\rvert\rvert_2=1}\lvert\lvert BA\mathbf{x}\rvert\rvert_2.
\end{equation}
SpectR improves routing accuracy at the cost of additional computation by using the entire $A^*$ matrix for routing decisions \citep{spectr}. We leverage both representations in LAG and perform adapter routing in two stages: Arrow-based retrieval and SpectR reranking. 

\subsection{Arrow Retrieval}

Once adapters are aligned, the new representation can be leveraged to perform routing without additional training or data access. For a specific LM layer $l$, there may be associated adapters from one or both of the adapter libraries $\mathcal{K}$ and $\mathcal{T}$. If both libraries apply, routing is performed separately for task adapters and knowledge adapters, and the selected adapter from each set is used. SpectR becomes infeasible with a large adapter library due to the memory requirements of storing and computing with the full matrices \citep{spectr}. We therefore perform an efficient first pass retrieval to select the $k$ adapters $\mathcal{A}$ producing the largest magnitude product between their associated arrow vector and the input vector $\mathbf{x}$:
\begin{equation}
    \mathcal{A} = \mathrm{topk}_i \lvert\mathbf{a}^*_i\mathbf{x} \rvert.
\end{equation}
The parameter $k$ can be chosen based on memory or time budgets, with a lower $k$ being more efficient and a higher $k$ reducing the impact of inaccurate arrow rankings.  

\subsection{SpectR Reranking}

\begin{figure}[t]
    \centering
    \includegraphics[width=\columnwidth]{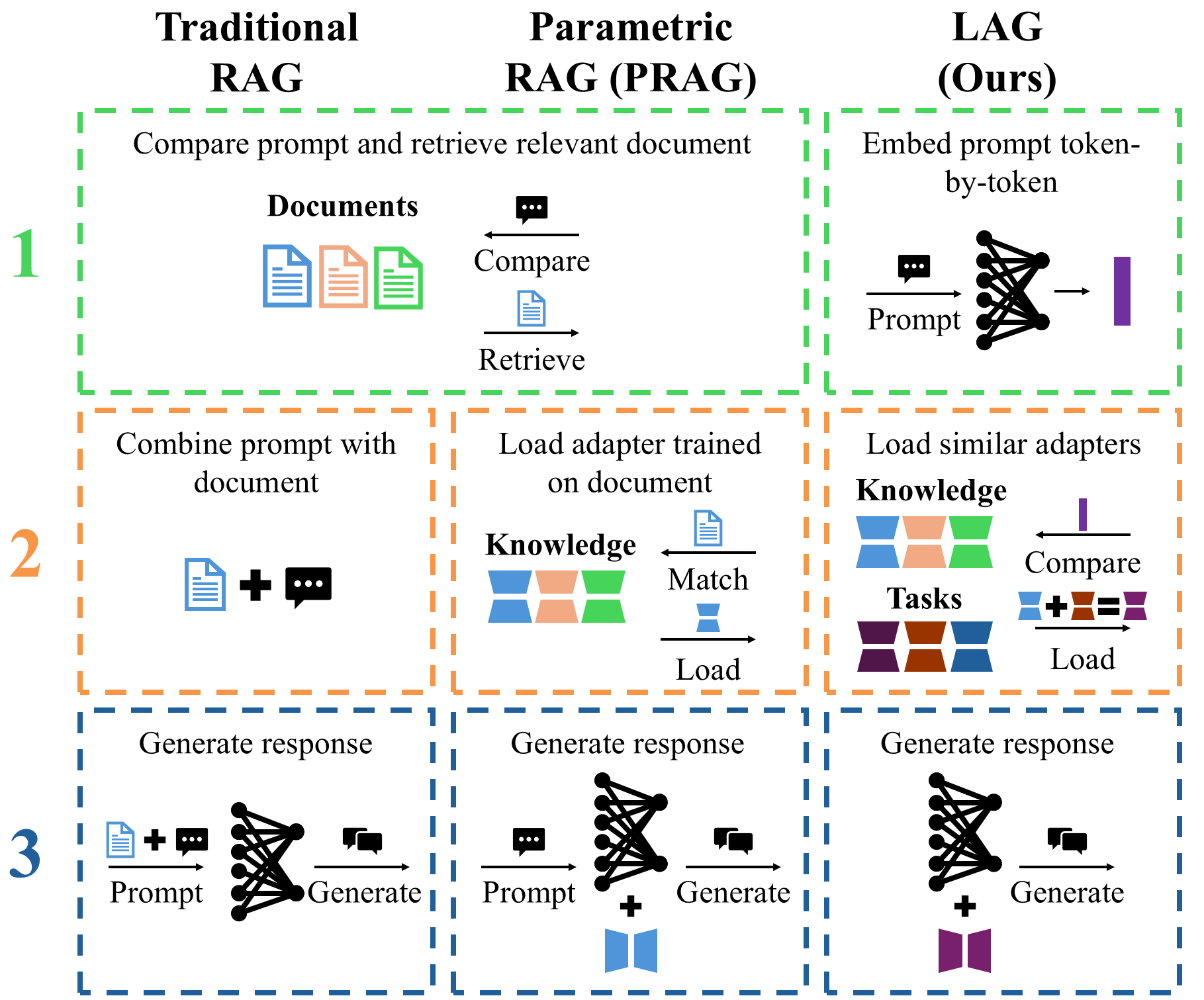}
    \caption{RAG and PRAG both identify the document most similar to the prompt. RAG includes the document in the prompt, while PRAG loads the corresponding adapter. LAG uses internal representations of the prompt to select and load knowledge and task adapters during generation.}
    \label{compare}
\end{figure}

We use SpectR scores on the subset of retrieved adapters $\mathcal{A}$, to more accurately compare and select the adapter most aligned with the input vector $\mathbf{x}$:
\begin{equation}
    \hat{A} = \mathrm{argmax}_{A^* \in \mathcal{A}}\lvert\lvert A^*\mathbf{x}\rvert\rvert_2.
    \label{eq:score}
\end{equation}
The selected adapter $(\hat{B}\text{, } \hat{A})$ is then used with the target layer weights $W_l$ to produce the output representation:
\begin{equation}
    \mathbf{h} = W_l\mathbf{x} + \hat{B}(\hat{A}\mathbf{x}),
\end{equation}
with the low-rank vector $\hat{A}\mathbf{x}$ precomputed and reused from Equation \ref{eq:score}. This two-stage process is used to select a single adapter for each position in the sequence. Figure \ref{compare} broadly compares LAG with the mechanisms of RAG and PRAG. 

\section{Experiments}

In this section, we experiment in practical scenarios where a large number of adapters are available for a diverse body of knowledge and tasks. We compare the theoretical requirements for each of the training and data-free approaches and confirm that SpectR is computationally infeasible for the large adapter library in our experiments. We explore our core setting with zero access to additional data or training, empirically demonstrating the benefits of our approach. We measure the impact of more aggressive filtering on LAG's performance, and compare LAG with alternative techniques for leveraging the data associated with our knowledge library.

% We experiment in practical scenarios where using SpectR is infeasible due to the computational requirements for considering a large number of adapters. We provide efficiency comparisons and explore our core setting with zero access to additional data or training. We then compare LAG with alternative approaches for leveraging the data associated with a knowledge library. 

\subsection{Data and Metrics} 

We leverage the KILT benchmark, a set of \emph{knowledge-intensive language task} datasets \citep{petroni-etal-2021-kilt}. KILT contains five tasks: fact checking, entity linking, slot filling, question answering (QA), and dialog generation (chat); all of which are grounded in knowledge from a shared collection of Wikipedia articles \citep{petroni-etal-2021-kilt}. No existing LoRAs correspond to these articles, so we must train a large adapter library ourselves. We choose a library of size $n=1000$, which is 2 orders of magnitude larger than previous SpectR experiments, yet still manageable to train on a single GPU. We filter the Wikipedia pages to the top 1000 referenced as provenance across the benchmark tasks. We use this set as our knowledge base and filter the task-specific datasets to only those examples grounded in these articles. Two datasets contain too few samples from the resulting selection and are dropped. Our final evaluation dataset is comprised of \textbf{fact checking:} FEVER \citep{Thorne18Fever}; \textbf{entity linking:} AIDA CoNLL-YAGO \citep{hoffart-etal-2011-robust}, WNED-WIKI, and WNED-CWEB \citep{wned}; \textbf{slot filling:} Zero Shot RE \citep{levy-etal-2017-zero} and T-REx \citep{elsahar-etal-2018-rex}; \textbf{QA}: Natural Questions \citep{kwiatkowski-etal-2019-natural} and TriviaQA \citep{joshi-etal-2017-triviaqa}; and \textbf{chat}: Wizard of Wikipedia \citep{wizard}. KILT prescribes different evaluation metrics depending on the dataset and task being evaluated \citep{petroni-etal-2021-kilt}. These include \textit{Accuracy} for tasks with a discrete output (fact checking, entity linking, and slot filling), \textit{Exact Match} for extractive QA, and \textit{F1 Score} for chat. We also introduce a normalized performance metric to more easily compare across tasks and to control for the differences in difficulty between datasets. Let $f_D(\mathrm{M})$ represent the score for the dataset-specific metric of model $\mathrm{M}$ evaluated on a dataset $D$. We produce a normalized task score $S_T$ using the performance of $\mathrm{M}$ compared to a strong reference model $\mathrm{R}$ across all datasets in the same task $T$:
\begin{equation}
    S_T = \sum_{D \in T}\frac{\lvert D \rvert}{\lvert T \rvert}\frac{f_D(\mathrm{M})}{f_D(\mathrm{R})},
\end{equation}
where $\lvert \cdot\rvert$ is the number of samples in $D$ or $T$. $S_T$ represents the average percentage of the reference model performance on task $T$ achieved by the model under evaluation. We refer to our reference model as the \textit{Oracle} model, which is the LM augmented with the ground-truth knowledge and task adapters for each query. For example, when answering a question about the first U.S. President, the Oracle model would use the \textit{QA} and \textit{George Washington} LoRA adapters. Table \ref{tab:metrics} includes a summary of the datasets used with their associated task, metric, and number of samples.

\begin{table}[t]
    \centering
    \small
    \begin{tabular}{cccr}
    Dataset & Task & Metric & Size\\
    \toprule
    WoW & Chat & F1 & 28546\\
    \midrule
    FEV & Fact & Acc & 24886\\
    \midrule
    AY2 & Link & Acc & 9056\\
    WnCW & Link & Acc & 995\\
    WnWi & Link & Acc & 237\\
    \midrule
    zsRE & Slot & Acc & 116\\
    T-REx & Slot & Acc & 280\\
    \midrule
    NQ & QA & EM & 3855\\
    TQA & QA & EM & 4583\\
    \end{tabular}
\caption{Filtered dataset information organized by task.}
\label{tab:metrics}

\end{table}

\subsection{Model and Adapter Libraries}

We use the Llama-3.2-3B-Instruct model as the LM in our experiments due to its generally good performance and efficient size amenable to running experiments using 1000s of LoRAs on a single GPU \citep{llama}.  

Evaluating LAG requires both task and knowledge LoRA libraries corresponding to our evaluation data. We therefore construct libraries with a LoRA adapter for each Wikipedia article and task. No hyperparameter tuning is performed since LAG is designed to work with externally sourced adapters. See Appendix for training and adapter details.

\subsubsection{Task Library} For each of the original five tasks, we fit a task-specific adapter using samples with provenance outside of our selected Wikipedia articles. The training data contains only the prompt and answer, not the corresponding articles. 

\subsubsection{Knowledge Library} We use synthetic continued pretraining to fit our knowledge adapters in a task-agnostic manner \citep{scpt}. We train one adapter per Wikipedia article. The KILT representation of each article includes a set of \textit{anchor} elements corresponding to hyperlinked entities contained in the text \citep{petroni-etal-2021-kilt}. The base LM is prompted to rewrite each article once per entity, emphasizing the focused entity in relation to the article contents. The original article is combined with these synthetically generated documents to increase the amount of training data used to fit each LoRA adapter. We train our knowledge adapters using the pretrained version of the LM to mitigate negatively impacting the existing instruction-tuning \citep{readapt}. The pretrained model is only used during LoRA training, and the adapters are applied to the instruction-tuned model during evaluation.

\subsection{Theoretical Efficiency}

Before empirically demonstrating the benefits of our approach, we discuss the expected efficiency gains of LAG in terms of required disk space, GPU memory, and computation. We summarize these comparisons in Table \ref{tab:computation}.

\begin{table}[t]
    \centering
    \small
    \begin{tabular}{rccc}
        & Arrow & SpectR & LAG \\
        \toprule
        Disk & $2nhr + nh$ & $2nhr$ & $2nhr$\\
        GPU & $2khr+nh$ & $2nhr$ & $2khr + nh$\\
         FLOPs & $2nh$ & $2nhr$ & $2h(n+rk)$\\
    \end{tabular}
    \caption{Overview of approximate FLOPs and best-case storage requirements in terms of parameters needed on disk and GPU assuming a library of $n$ rank-$r$ adapters with a hidden dimension of $h$ and top-$k$ filtering. LAG inherits the best between Arrow and SpectR for storage and is on par with Arrow in terms of computational efficiency.}
    \label{tab:computation}
\end{table}

\subsubsection{Disk Space} Given a linear layer with input and output dimension $h$ and a library of $n$ adapters with rank $r$: Arrow, SpectR, and LAG all require the storage of $2nhr$ LoRA parameters. Each of the $n$ adapters is composed of two $h \times r$ matrices. Arrow routing uses adapters in their original form and requires an additional $nh$ parameters to store the arrow vector from each adapter in the library \citep{arrow}. LAG also uses arrow vectors, but because the adapters are stored in their aligned representation, the arrow vectors are captured by the first row of the $A^*$ matrix from Equation \ref{Astar}, preventing the need for extra storage. Like LAG, SpectR only requires the aligned adapter library as it performs routing using the $A^*$ matrix instead of using arrow vectors \citep{spectr}.

\subsubsection{GPU Memory} Depending on available memory, all the parameters can be loaded onto the GPU to eliminate data transfer. If memory is limited, Arrow and LAG can take advantage of their ability to filter down to only $k<<n$ adapters per token using arrow vectors and load only the selected LoRAs into memory. For a token sequence of length $s$, the $nh$ arrow parameters are used to choose the $sk$ adapters for the sequence \citep{arrow}. In the worst case, all $n$ adapters are required in memory if the sequence is longer than $n/k$ tokens and positions in the sequence share little overlap in the relevant knowledge or tasks needed. However, \citet{spectr} demonstrated the intuition that sequences do result in high overlap, which reduces the number of adapters needed in memory. In the best case, the same $k$ adapters would be selected for all tokens, resulting in a GPU storage requirement of $nh + 2khr$ parameters for both Arrow and LAG\footnote{E.g. 1 million rank-6 adapters with $k=20$ and $h=4096$ would require 49B extra parameters w/ SpectR versus 4B w/ LAG.}. In the unlikely worst case, the GPU requirements are the same as the disk space.

\subsubsection{Computation}

We compare the additional floating-point operations (FLOPs) required, assuming each method selects a single adapter per-token. SpectR requires $n$ matrix-vector products using the $A^*$ matrices of each adapter and an additional product using the $B^*$ matrix from the chosen adapter \citep{spectr}. This results in $2nhr + 2hr = 2hr(n+1) \approx 2nhr$ FLOPs. Arrow is the most efficient method, requiring only $2nh$ FLOPs to choose an adapter using the $n$ arrow vector dot products \citep{arrow}. Another $2hr$ FLOPs are used to multiply by each of the LoRA matrices of the chosen adapter, incurring a total of $2nh + 2hr + 2hr = 2h(n + 2r) \approx 2nh$ FLOPs. LAG inherits the same $2nh$ FLOPs from Arrow to choose the top-$k$ adapters and then requires $2khr$ more to process those $k$ LoRAs using SpectR, a total of $2nh + 2khr = 2h(n+rk)$ FLOPs. In our experiments, $rk << n$, yielding computation requirements for LAG similar to Arrow. Next, we use these efficiency gains for our experiments in scenarios where the LoRA library is too large for SpectR.

\subsection{Data and Training Free}

\begin{table}[t]
    \centering
    \small
    \begin{tabular}{r|ccc|c}
        & Instr & Arrow & LAG & Oracle\\
        \toprule
        WoW & 12.5 & \textbf{17.9} & 17.7 & 18.1\\
        \midrule
        FEV & 66.0 & 89.2 & \textbf{89.6} & 90.2\\
        \midrule
        AY2 & 21.0 & 45.2 & \textbf{62.4} & 68.2\\
        WnCw & 15.8 & 23.1 & \textbf{47.2} & 67.6\\
        WnWi & 13.5 & 32.5 & \textbf{61.2} & 73.0\\
        \midrule
        zsRE & 24.1 & 42.2 & \textbf{44.8} & 49.1\\
        T-REx & 18.9 & 45.4 & \textbf{46.1} & 53.6\\
        \midrule
        NQ & 27.0 & 26.1 & \textbf{30.9} & 38.8\\
        TQA & \textbf{50.8} & 44.2 & 46.5 & 50.8\\

    \end{tabular}
    \caption{LAG outperforms other data and training-free approaches for the majority of the evaluated datasets.}
    \label{tab:all}
\end{table}

\begin{table}
    \centering
    \small
    \begin{tabular}{r|ccccc|c}
    & Chat & Fact & Link & Slot & QA & AVG\\
    \toprule
    Instr & 69.3 & 73.2 & 29.8 & 39.4 & \textbf{86.0} & 59.5\\
    
    Arrow & \textbf{99.2} & 99.0 & 62.6 & 85.0 & 78.0 & 84.8\\
    
    LAG & 98.2 & \textbf{99.4} & \textbf{89.2} & \textbf{87.5} & \textbf{86.0} & \textbf{92.1}\\
    \end{tabular}
    \caption{LAG produces an average gain of 7 points in normalized performance over Arrow across all tasks.}
    \label{tab:both}
\end{table}

We explore our core objective of leveraging large LoRA libraries in a scenario where access to the corresponding adapter data or additional training for learning to route is unavailable. Following Table \ref{tab:computation}, SpectR is roughly $8\times$ more expensive per-token than Arrow and LAG using the rank-8 LoRAs in our library. Indeed, we found SpectR to be intractable for our evaluation using our library of 1000 adapters, exactly the issue LAG was developed to address. Therefore, we first compare LAG against Arrow routing and the instruction-tuned model without any adapters. We also report the performance of the Oracle model, using the ground-truth knowledge and task adapters applied to each sample. The Oracle provides a reference for the model's achievable performance. For LAG, we use $k=20$ for filtering, which reduces the number of adapters considered by SpectR by 98\%. We later explore the impact of $k$ on our results. We select a single adapter to apply per-token and per-layer for both Arrow and LAG. 

Table \ref{tab:all} displays the result for each dataset using the associated KILT metric. The increase in performance when using the Oracle model provides insight into how much the knowledge and task adapters improve performance over the instruct model. The Oracle model performs much better than the base instruct model across all tasks except for QA, where the Oracle performs better on NQ but achieves the same performance on TriviaQA. This could indicate that these datasets were highly represented in the original LM training data or that the instruction-tuning is well-suited for QA in general. LAG consistently outperformed both baselines. The two exceptions were the chat task, where Arrow and LAG both do almost as well as the Oracle model, and on TriviaQA where the Oracle performance suggests the LoRA adapters provide little value. Table \ref{tab:both} provides the normalized performance per-task. LAG captures 92.1\% of the Oracle performance on average, scoring 7.3 points higher than Arrow and improving the instruct model by 32.6 points. 

\subsection{Aggressive Filtering}

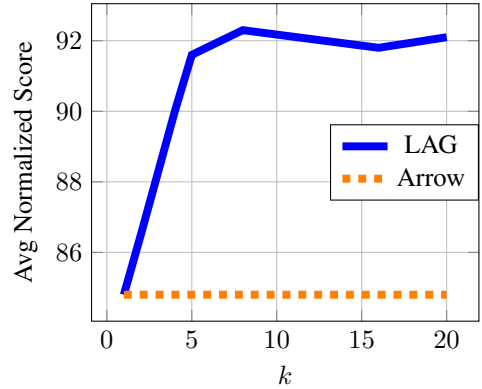
\begin{figure}
    \centering
    \begin{tikzpicture}
    \begin{axis}[
    width=.8\columnwidth,
	xlabel=$k$,
	ylabel=Avg Normalized Score,
	grid=both,
        legend style={at={(1,.5)}, anchor=east},
	no marks]
\addplot[line width=3pt,solid,color=blue] %
	table[x=k,y=avg,col sep=comma]{results/k.csv};
\addlegendentry{LAG};
\addplot[line width=3pt,dashed,color=orange] 
	table[x=k,y=avg,col sep=comma]{results/arrow.csv};
\addlegendentry{Arrow};
\end{axis}
\end{tikzpicture}
\caption{The average normalized performance of LAG goes up and plateaus as $k$ increases. LAG is equivalent to Arrow routing at $k=1$ and would be equivalent to SpectR if $k$ could be set to the total number of adapters in the library.}
\label{fig:kplot}
\end{figure}

We filtered the LoRAs using $k=20$ with LAG because using SpectR on the entire adapter library is intractable. The value of $k$ modulates the trade-off between Arrow efficiency and the superior accuracy of SpectR at a higher computational cost. With $k=1$, LAG becomes equivalent to Arrow routing because the SpectR step is forced to choose the given adapter. As $k$ increases, there are better chances for Arrow to include the appropriate adapters for SpectR reranking. For example, \citet{spectr} showed Arrow top-$k$ accuracy increasing by more than $3\times$ when going from top-1 to top-4 in their experiments. If $k$ is set to the total number of LoRAs in the library, then LAG is equivalent to SpectR, as no Arrow filtering occurs. 

Figure \ref{fig:kplot} shows the average normalized performance of LAG across tasks as a function of $k$. The performance rises steeply and then flattens around $k=5$, suggesting that more aggressive Arrow filtering improves efficiency with minimal sacrifice to downstream performance. In practice, the value of $k$ can be chosen for the given compute budget, or future research could explore setting $k$ dynamically as a form of test-time scaling \citep{testtime}.

\subsection{Knowledge Access}

Finally, we loosen our core objective and explore LAG in the case where the documents associated with the knowledge library are available during inference. This scenario allows for a comparison with alternative methods of incorporating knowledge into the LM, such as Retrieval-Augmented Generation (RAG) and Parametric RAG (PRAG) \citep{rag, prag}. Specifically, we use documents relevant to the query to either augment the prompt (RAG) or to retrieve the adapter trained on the selected document (PRAG). In keeping with \citet{prag}, we use BM25 \citep{bm25} for document retrieval in both cases. We continue to use $k=20$ for LAG to more easily compare with our previous results.

We include PRAG and RAG as individual baselines using each approach to augment the instruction-tuned model with additional knowledge. We then evaluate combinations using LAG, PRAG, or RAG for knowledge, with LAG selecting the task adapters in all cases. The individual dataset performances are shown in Table \ref{tab:other}. All LAG combinations perform better than either PRAG or RAG alone, except on QA where only RAG + LAG outperforms. The best method for incorporating knowledge varied across tasks, with no statistically significant winner. Table \ref{tab:otherc} reports the normalized task scores where using LAG for both knowledge and tasks performed best on fact checking and entity linking, while PRAG-LAG did best on chat, and RAG-LAG on slot filling and QA. Notably, RAG-LAG achieved a 102.7 normalized score on slot filling, meaning it outperformed the Oracle model by 2.7\%. The samples for slot filling are of the form `\emph{entity} [SEP] \emph{relationship}', making BM25 especially effective since the entity and relationship can both be found in the document containing the necessary information. This contrasts with a task like entity linking, where samples can contain arbitrary content with a single entity marked for identification. The majority of the content in each sample can be unrelated to the correct response. LAG's per-token selection of adapters provides more flexibility in such cases, and LAG did achieve a better score on entity linking in our evaluation.

\begin{table}[t]
    \centering
    \small
    \begin{tabular}{r|cc|ccc}
        & PRAG & RAG & LAG & P-LAG & R-LAG\\
        \toprule
        WoW & 12.7 & 12.6 & 17.7 & \textbf{17.9} & 17.8\\
        \midrule
        FEV & 68.4 & 78.6 & \textbf{89.6} & 89.5 & 89.2\\
        \midrule
        AY2 & 24.6 & 23.4 & \textbf{62.4} & 61.8 & 57.2\\
        WnCw & 20.6 & 16.9 & \textbf{47.2} & 44.9 & 32.9\\
        WnWi & 19.4 & 21.1 & 61.2 & \textbf{62.4} & 47.3\\
        \midrule
        zsRE & 29.3 & 42.2 & 44.8 & 44.8 & \textbf{47.4}\\
        T-REx & 25.4 & 41.4 & 46.1 & 46.4 & \textbf{56.4}\\
        \midrule
        NQ & 24.7 & 32.6 & 30.9 & 30.8 & \textbf{35.9}\\
        TQA & 48.1 & 48.2 & 46.5 & 47.0 & \textbf{49.0}\\

    \end{tabular}
    \caption{Using LAG to incorporate task-capabilities outperforms PRAG or RAG alone, with variation across datasets. P-LAG: PRAG + LAG, R-LAG: RAG + LAG.}
    \label{tab:other}
\end{table}

\begin{table}
    \centering
    \small
    \begin{tabular}{r|ccccc|c}
    & Chat & Fact & Link & Slot & QA & AVG\\
    \toprule
    PRAG & 70.2 & 75.9 & 35.3 & 50.9 & 80.5 & 62.6\\
    RAG & 69.9 & 87.2 & 33.2 & 79.9 & 89.9 & 72.0\\
    \midrule
    LAG & 98.2 & \textbf{99.4} & \textbf{89.2} & 87.5 & 86.0 & 92.1\\
    P-LAG & \textbf{99.0} & 99.3 & 88.2 & 88.0 & 86.4 & 92.2\\
    R-LAG & 98.6 & 98.9 & 80.0 & \textbf{102.7} & \textbf{94.7} & \textbf{95.0}\\
    \end{tabular}
    \caption{Normalized performance across tasks. Different LAG combinations perform better across the tasks. Combining RAG with LAG results in the best average performance.}
    \label{tab:otherc}
\end{table}

% \section{Discussion}

% Our problem setting focuses on data and training-free approaches for adapter selection and combination. 

\section{Conclusion}

We introduce LoRA-Augmented Generation (LAG), a completely unsupervised approach for filtering a large LoRA library and applying the most suitable adapters at test time on a per-token and per-layer basis. LAG addresses scalability issues with previous solutions while maintaining superior downstream task performance over alternative training and data-free approaches. We evaluated LAG on a set of multiple knowledge-intensive tasks, including fact checking, entity linking, slot filling, question answering, and chat. We demonstrated LAG's ability to select from this diverse set of knowledge and task capabilities and efficiently incorporate the selected adapters into the LM. LAG significantly outperformed Arrow routing with access to the same adapter libraries. 
% We explored the sensitivity of performance to the aggressiveness of LAG filtering and discussed the equivalence of LAG to Arrow and SpectR at the extremes. 
We demonstrated that more aggressive arrow filtering can heavily reduce necessary computation with minimal degradation in task performance. We performed additional experimentation using the Wikipedia articles corresponding to our knowledge library. This allowed us to evaluate methods such as Retrieval-Augmented Generation (RAG) and Parametric RAG, and to combine these methods with LAG. Using LAG to incorporate task capabilities significantly outperformed using RAG or PRAG alone, but there was not a statistically significant difference between the combined approaches. The performance varied by task, and we discussed how different tasks might have characteristics that make each of the various approaches for incorporating knowledge with LAG more suitable in certain circumstances.

Overall, LAG successfully unifies the existing approaches in the area of unsupervised adapter routing, incorporating the efficiency of Arrow and the discriminative power of SpectR to provide scalable multi-task performance. 

\bibliography{custom, anthology}

\appendix

\section{Adapter Details}
\label{details}

We fit LoRA adapters for our knowledge and task libraries using the \emph{peft} package \citep{peft}. We use common hyperparameter settings, purposely not optimizing for LAG, which is designed to work with externally trained adapters. We use a learning rate of $1e^{-4}$ and LoRA dropout of $0.05$ for both libraries. Following \citet{spectr}, we use rank-8 LoRAs for our task adapters, targeting the k\_proj, q\_proj, v\_proj, and o\_proj attention layers of the instruction-tuned model. We train for a single epoch with a batch size of 8. We use rank-6 knowledge adapters targeting the gate\_proj, up\_proj, and down\_proj layers, following work suggesting that transformers store knowledge in their feed-forward layers \citep{geva-etal-2021-transformer}. These are fit using the pretrained version of the model to mitigate impacting the instruction-following capabilities \citep{readapt}. We train these LoRAs for 4 epochs with a batch size of 1. For both sets of adapters, we use a LoRA $\alpha$ which is twice the rank. All training and inference was done using a single Nvidia A100 GPU.

\end{document}